\documentclass[10pt,twocolumn,letterpaper]{article}

\usepackage[pagenumbers]{cvpr} 
\usepackage{multirow} 

\usepackage[dvipsnames]{xcolor}

\definecolor{cvprblue}{rgb}{0.21,0.49,0.74}
\usepackage[pagebackref,breaklinks,colorlinks,citecolor=cvprblue]{hyperref}

\title{Rethinking the Elementary Function Fusion for Single-Image Dehazing}

\author{
  Yesian Rohn$^{1}$ \\
  \and
  $^{1)}$ Shanghai Key Lab of Intelligent Information Processing, School of Computer Science, Fudan University \\
  \and
  \texttt{\{xsye20\}@fudan.edu.cn}
}

\begin{document}

\maketitle

\begin{abstract}

This paper addresses the limitations of physical models in the current field of image dehazing by proposing an innovative dehazing network—CL2S. Building on the DM2F model, it identifies issues in its ablation experiments and replaces the original logarithmic function model with a trigonometric (sine) model. This substitution aims to better fit the complex and variable distribution of haze. The approach also integrates the atmospheric scattering model and other elementary functions to enhance dehazing performance. Experimental results demonstrate that CL2S achieves outstanding performance on multiple dehazing datasets, particularly in maintaining image details and color authenticity. Additionally, systematic ablation experiments supplementing DM2F validate the concerns raised about DM2F and confirm the necessity and effectiveness of the functional components in the proposed CL2S model. Our code is available at \url{https://github.com/YesianRohn/CL2S}, where the corresponding pre-trained models can also be accessed.

\end{abstract}    
\section{Introduction}

\subsection{Background and Significance}

In recent years, with the acceleration of industrialization and urban expansion, environmental problems have become increasingly prominent, with haze becoming a severe challenge faced by many cities. Haze not only poses a threat to human health but also significantly affects daily life and socio-economic activities. Image degradation caused by haze, such as color distortion, reduced contrast, and blurred details, severely hinders normal operations and efficiency improvements in various fields. For example, in autonomous driving scenarios, reduced visibility due to fog significantly increases driving risks; in satellite remote sensing, haze obscures the true state of the ground, affecting the accuracy of data analysis.

Therefore, developing efficient and precise image dehazing techniques is not only about restoring the visual aesthetics of images but also ensuring the stable operation and reliable decision-making of various vision-dependent technological systems. This requires algorithms to effectively remove visual interference caused by haze, restore the true color and details of images, and maintain robustness under complex and variable environmental conditions, minimizing additional distortions such as over-smoothing, structural distortion, or color distortion introduced by dehazing processing.

Currently, image dehazing technology is transitioning from traditional physical model-based methods to deep learning-driven strategies. The former attempts to reverse estimate and remove atmospheric scattering effects by establishing optical scattering models, emphasizing theoretical foundations and physical realism; the latter leverages the powerful learning capabilities of deep neural networks to learn dehazing mappings directly from large amounts of paired hazy-clear images, focusing on practical effects and generalization ability. Each method has its advantages and limitations. While deep learning methods often outperform in terms of effectiveness, they require higher computational resources and have poorer model interpretability. Therefore, in-depth research and integration of the advantages of both methods, and innovation to address the shortcomings of existing technologies, have become urgent problems in this field.

\subsection{Challenges in Dehazing}

\subsubsection{\textbf{Physical Models and Prior Knowledge}}

The atmospheric scattering model, a cornerstone of traditional dehazing methods, inherently suffers from the ill-posed problem of solving coupled equations for atmospheric light intensity and atmospheric transmission functions. Consequently, researchers usually rely on image priors, such as the dark channel prior, to constrain the solution space of the model. Although these priors are highly regarded for their simplicity and effectiveness in revealing the statistical properties of natural images, especially their successful application in dehazing, their general applicability in complex real-world environments is limited. They often require manual parameter adjustments for each image, limiting the model's generalization ability and level of automation.

\subsubsection{\textbf{Deep Learning and Dataset Bias}}

The introduction of deep learning technology provides new solutions for image dehazing, with its learnable capabilities effectively enhancing the upper limit of dehazing performance. However, the inherent distribution differences between supervised training on synthetic data and actual hazy images make models difficult to apply directly to real-world scenarios. Moreover, deep learning models lacking direct physical model constraints may produce unexpected results. While many supervised dehazing methods are equally effective in other image restoration tasks, this more reflects the generalization ability of the model rather than specific design for haze models. Therefore, unsupervised, semi-supervised, and zero-shot learning strategies, as ways to reduce reliance on paired data, remain to be explored.

\subsubsection{\textbf{Inhomogeneous Haze Distribution and \\ Feature Channel Differences}}

The uneven distribution of haze in images and the differences in haze representation across different feature channels pose higher requirements for dehazing algorithms. Most current deep learning algorithms apply uniform weights to all spatial pixels and feature channels, failing to fully consider haze density variations and inter-channel characteristic differences, which limits the optimization of dehazing effects. An ideal dehazing algorithm should automatically adapt to different haze density regions and apply optimal attention to each feature channel accordingly.

\subsection{Related Work and Research}

Traditional single-image dehazing methods are primarily based on the atmospheric scattering model~\cite{cantor1978optics}, focusing on designing handcrafted features such as the dark channel prior~\cite{5567108} and color attenuation prior~\cite{7128396}. However, these prior-based methods may not sufficiently represent complex scenes in practice, often generating artifacts in the results. Early learning-based methods~\cite{2016DehazeNet, Ren2016SingleID} used deep neural networks to predict the transmission map and atmospheric light in the physical model to obtain potential clear images. However, inaccuracies in estimation could accumulate, hindering reliable inference of haze-free images. Consequently, data-driven methods~\cite{Qin2019FFANetFF, deng2019deep, Chen_2021_CVPR, Guo2022ImageDT, Liu2021FromST} have rapidly developed. FFANet~\cite{Qin2019FFANetFF} introduced a feature attention module, using channel and pixel attention to improve haze removal effects. DeHamer~\cite{Guo2022ImageDT} combined CNN~\cite{726791} and Transformer~\cite{attention} for image dehazing, integrating the long-range attention of Transformers with the local attention of CNN features. It should be noted that these methods do not consider the physical properties in the haze formation process. The dehazing network DCPDN~\cite{2018Densely} considers the atmospheric scattering model, jointly learning the transmission map, atmospheric light, and haze-free image to capture their relationships. DM2F~\cite{deng2019deep} views the dehazing process as a hierarchical separation model, introducing additional auxiliary elementary functions to fit the physical model. However, DM2F merely stacks elementary functions without considering the completeness and rationality of the function selection.
\section{Method}

\begin{figure*}[htbp]
  \centering
  \includegraphics[width=\textwidth]{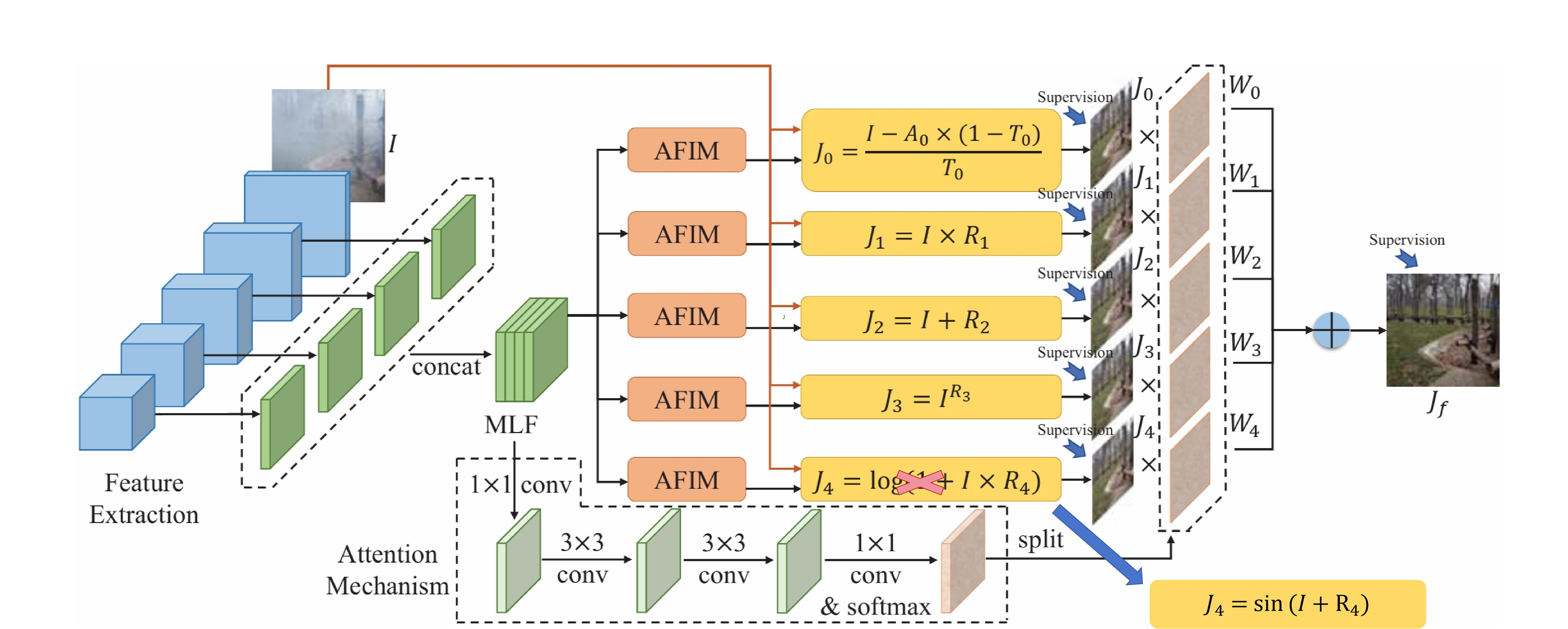} 
  \caption{CL2S architecture, with specific modifications replacing $J_{4}$ in the bottom right corner.}
  \label{fig:CL2S}
\end{figure*}

Figure \ref{fig:CL2S} illustrates our proposed network structure, named CL2S (Change Logarithmic to Sinusoidal function), which is designed by modifying the DM2F \cite{deng2019deep} model, replacing the original logarithmic function with a more effective trigonometric function. This improvement leverages the advantageous properties of trigonometric functions, making the atmospheric scattering model and a composite model of three necessary elementary functions composed of four operator models more suitable for dehazing tasks. Furthermore, we systematically weight and fuse the outputs of each dehazing component through the learned attention maps to generate the final dehazed image.

It is worth noting that despite the changes in the core component $J_{4}$, CL2S retains consistency with DM2F in many aspects, continuing to use the efficient ResNeXt \cite{Xie2016AggregatedRT} model as the cornerstone for feature extraction and preserving the attention feature aggregation module to optimize feature weight allocation strategies. Regarding the selection of elementary functions, we reconsidered the original design of DM2F, clearly identifying the limitations of the existing function configuration. In the subsequent ablation experiments, we thoroughly present our theoretical basis and empirical analysis, further demonstrating the effectiveness and rationality of the CL2S design.

\subsection{Review of Baseline Method}
DM2F first utilizes the attention feature integration module to generate atmospheric light and medium-integrated features, then learns the necessary parameters for the atmospheric scattering model:

\[ J_0(p) = I(p) - A_0 \times (1 - T_0(p)) \]

where \(J_0\) represents the dehazed result, \(I\) is the input hazy image, \(A_0\) is the estimated atmospheric light, and \(T_0\) is the transmission map, estimated by applying convolution and activation layers on the atmospheric features.

Furthermore, to capture more diverse dehazing information, four different operator models are introduced, each based on operations of different elementary functions: linear functions (addition, multiplication), exponential function, and logarithmic function, to decompose the imaging principle and predict the dehazed results \(J_1, J_2, J_3, J_4\) respectively.
The formulas for each operator model are as follows:
\\ 
\indent \textbf{Multiplication Model}
\[ J_1(p) = I(p) \times R_1(p) \]

\textbf{Addition Model}
\[ J_2(p) = I(p) + R_2(p) \]

\textbf{Exponential Model}
\[ J_3(p) = (I(p))^{\left(R_3(p)\right)} \]

\textbf{Logarithmic Model}
\[ J_4(p) = \log(1 + I(p) \times R_4(p)) \]

where \(R_i\) corresponds to the feature learning layer in each model, obtained by combining and allocating attention to the hierarchical features. After obtaining the predictions of different dehazing models, an attention mechanism is used to integrate these predictions to get the final network output. To this end, DM2F learns five attention maps from the multi-layer integrated features by executing a 1×1 convolution layer, two 3×3 convolution layers, another 1×1 convolution layer, and a softmax layer. Then, the final result (denoted as $J_f$) is calculated as follows, where \(W_0, W_1, W_2, W_3, W_4\) are the learned attention map weights for the dehazing results \(J_0, J_1, J_2, J_3, J_4\):
\[ J_f = W_0 \cdot J_0 + W_1 \cdot J_1 + W_2 \cdot J_2 + W_3 \cdot J_3 + W_4 \cdot J_4 \]

\subsection{Elementary Function Fitting}

In exploring the scope of elementary function models, logarithmic and exponential operations as nonlinear transformation components show functional overlap due to their continuity and the characteristic of mapping their value range to the [0,1] interval. However, beyond these two, trigonometric functions, especially the sine function, constitute an underexplored potential area. Inspired by the Transformer model~\cite{attention} in utilizing positional encoding, we hypothesize that incorporating trigonometric operations into the dehazing framework can effectively enhance the model's adaptability and robustness in complex scenes.

Specifically, we introduce a sine model as a new dehazing component, mathematically defined as:
\[ J_4(p) = \sin{(I(p) + R_4(p))} \]
Here, $I(p)$ represents the original image intensity at pixel position $p$, and $R_4(p)$ is a specific correction term at that position, aiming to adjust the phase of the sine wave and capture potential periodic haze distribution features.

The final dehazed output result comprehensively considers the contributions of all elementary function models and is derived through weighted fusion, expressed as:
\[ J_f = W_0 \cdot J_0 + W_1 \cdot J_1 + W_2 \cdot J_2 + W_3 \cdot J_3 + W_4 \cdot J_4 \]

\section{Experimental Results and Analysis}

\begin{table*}[htbp]
\centering
\caption{Comparisons on some dehazing datasets.}
\begin{tabular}{c|cc|cc|cc}
\hline
\multirow{2}{*}{Method} & \multicolumn{2}{c|}{O-HAZE~\cite{Ancuti2018OHAZEAD}} & \multicolumn{2}{c|}{HazeRD~\cite{8296874}} & \multicolumn{2}{c}{RESIDE~\cite{Ren-CVPR-2018}} \\ \cline{2-7} 
                        & \multicolumn{1}{c|}{PSNR$\uparrow$} & SSIM~\cite{1284395}$\uparrow$  & \multicolumn{1}{c|}{CIEDE2000~\cite{CIEDE2000}$\downarrow$} & SSIM$\uparrow$  & \multicolumn{1}{c|}{PSNR$\uparrow$} & SSIM$\uparrow$  \\ \hline
DCP ~\cite{5567108}          & \multicolumn{1}{c|}{16.59} & 0.735 & \multicolumn{1}{c|}{17.9014}   & 0.534 & \multicolumn{1}{c|}{16.62} & 0.8179 \\ \hline
NLD  ~\cite{7780554}       & \multicolumn{1}{c|}{16.61} & 0.750 & \multicolumn{1}{c|}{16.4010}   & 0.577 & \multicolumn{1}{c|}{17.27} & 0.7500 \\ \hline
MSCNN ~\cite{mscnn}       & \multicolumn{1}{c|}{19.07} & 0.765 & \multicolumn{1}{c|}{13.7952}   & 0.624 & \multicolumn{1}{c|}{17.57} & 0.8100 \\ \hline
DehazeNet ~\cite{2016DehazeNet}     & \multicolumn{1}{c|}{16.21} & 0.666 & \multicolumn{1}{c|}{17.1261}   & 0.479 & \multicolumn{1}{c|}{21.14} & 0.8500 \\ \hline
DCPDN ~\cite{2018Densely}         & \multicolumn{1}{c|}{22.78} & 0.742 & \multicolumn{1}{c|}{14.6251}   & 0.546 & \multicolumn{1}{c|}{28.13} & 0.9592 \\ \hline
GFN ~\cite{Ren-CVPR-2018}           & \multicolumn{1}{c|}{22.58} & 0.737 & \multicolumn{1}{c|}{16.3619}   & 0.511 & \multicolumn{1}{c|}{22.30} & 0.8800 \\ \hline
PDNet ~\cite{pdnet}         & \multicolumn{1}{c|}{17.40} & 0.658 & \multicolumn{1}{c|}{16.9360}   & 0.495 & \multicolumn{1}{c|}{22.83} & 0.9210 \\ \hline
AOD-Net ~\cite{2017AOD}       & \multicolumn{1}{c|}{19.59} & 0.679 & \multicolumn{1}{c|}{16.6743}   & 0.500 & \multicolumn{1}{c|}{20.86} & 0.8788 \\ \hline
DM2F (Baseline-Paper)  ~\cite{deng2019deep}     & \multicolumn{1}{c|}{\textbf{25.19}} & \textbf{0.777} & \multicolumn{1}{c|}{12.9285}   & 0.656 & \multicolumn{1}{c|}{34.29} & \textbf{0.9844} \\ \hline
\hline
DM2F (Baseline-Code)  ~\cite{deng2019deep}     & \multicolumn{1}{c|}{24.41} & 0.761 & \multicolumn{1}{c|}{11.5856}   & \textbf{0.669} & \multicolumn{1}{c|}{34.99} & 0.9804 \\ \hline
CL2S (Ours)      & \multicolumn{1}{c|}{\textbf{24.58}} & \textbf{0.763} & \multicolumn{1}{c|}{\textbf{11.4193}}   & 0.667 & \multicolumn{1}{c|}{\textbf{35.36}} & \textbf{0.9808} \\ \hline
\end{tabular}
\end{table*}

\begin{figure*}[htbp]
  \centering
  \includegraphics[width=0.9\textwidth]{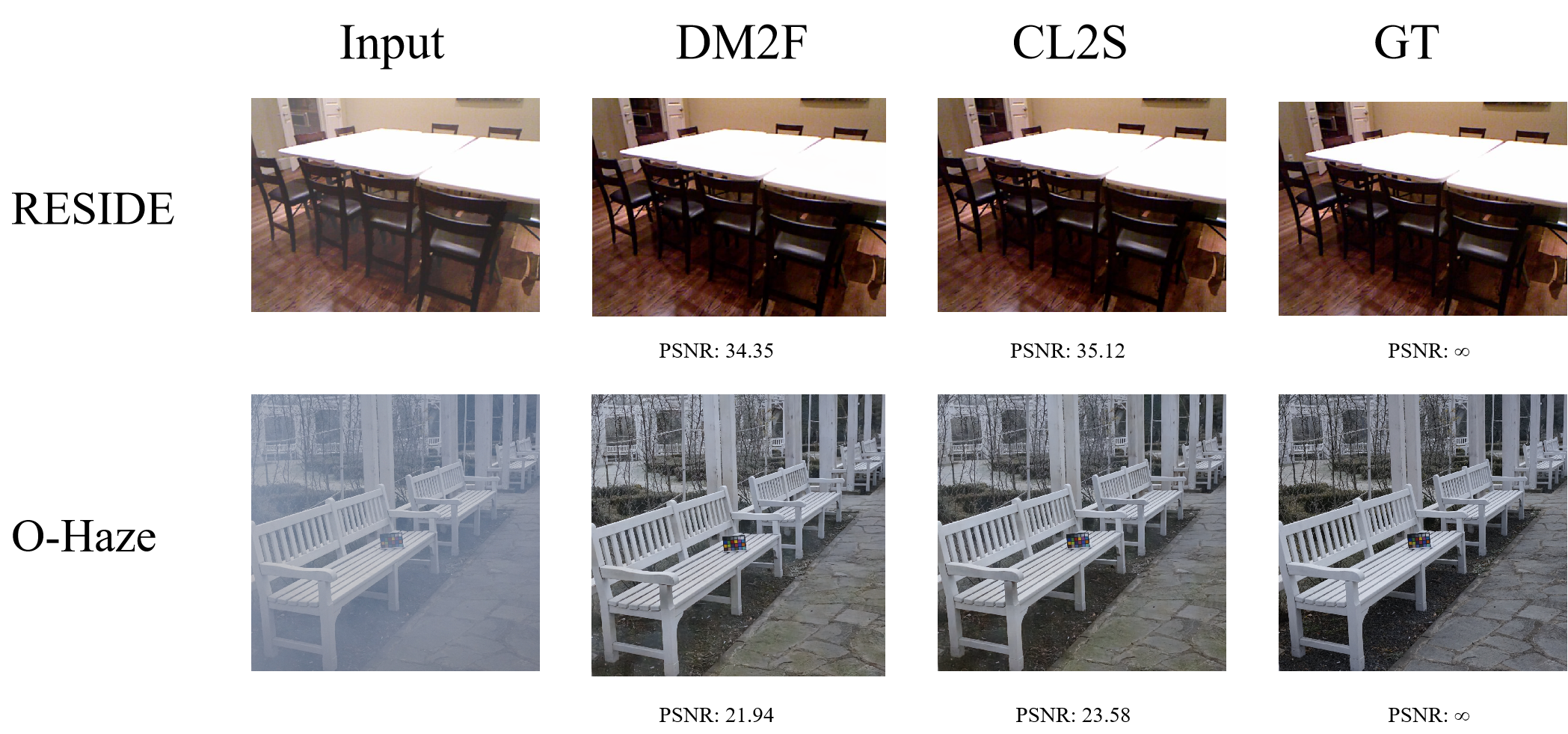} 
  \caption{Examples of CL2S and DM2F performance on mainstream datasets}
  \label{fig:bk}
\end{figure*}

\begin{figure*}[tbp]
  \centering
  \includegraphics[width=0.8\textwidth]{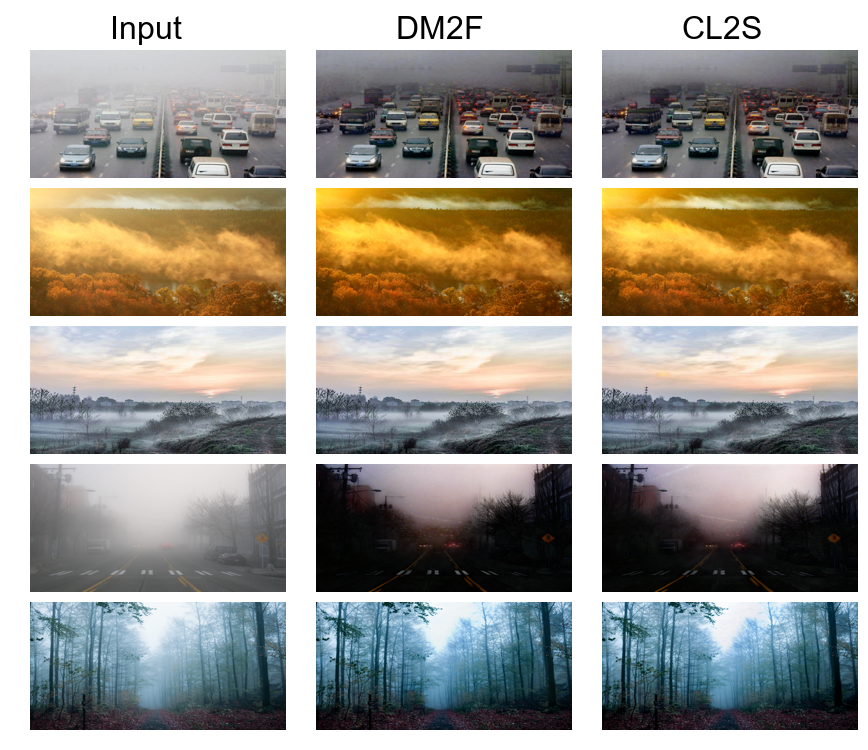} 
  \caption{Performance of CL2S and DM2F on C-Haze}
  \label{fig:comparison_table}
\end{figure*}

We compared the performance of our dehazing network against baseline models and other mentioned methods, including DCP~\cite{5567108}, NLD~\cite{7780554}, MSCNN~\cite{mscnn}, DehazeNet~\cite{2016DehazeNet}, AOD-Net~\cite{2017AOD}, GFN~\cite{Ren-CVPR-2018}, DCPDN~\cite{2018Densely}, PDNet~\cite{pdnet}, and the baseline method DM2F~\cite{deng2019deep}. For quantitative comparisons, we employed three widely used evaluation metrics: Peak Signal-to-Noise Ratio (PSNR), Structural Similarity Index (SSIM)~\cite{1284395}, and CIEDE2000~\cite{CIEDE2000}. Figures \ref{fig:bk} and \ref{fig:comparison_table} showcase the performance of our method and the baseline model on benchmark dataset samples and our collected C-Haze dataset.

\subsection{Datasets}
In this paper, we adopted a series of widely recognized datasets to systematically validate the effectiveness of the proposed dehazing algorithm CL2S. These datasets include RESIDE~\cite{Ren-CVPR-2018}, O-HAZE~\cite{Ancuti2018OHAZEAD}, HazeRD~\cite{8296874}, and our own collected five hazy images (C-Haze) for visualization analysis, to comprehensively cover image dehazing challenges under different scenes and haze concentration conditions.

\textbf{RESIDE:} Following the benchmark methods, we processed the RESIDE dataset by applying ITS for training and using SOTS for evaluation. We also focused on testing the model with other smaller datasets, specifically including HazeRD and our carefully collected C-Haze dataset. Notably, we only used the RESIDE dataset during training to ensure the model's adaptability to a wider range of scenes.

\textbf{O-HAZE:} The O-HAZE dataset consists of hazy and corresponding haze-free images of 45 outdoor scenes, capturing visual information under different haze concentrations. According to the competition regulations, we selected 35 image pairs for training and the remaining 10 pairs for testing to ensure the stability and accuracy of the algorithm.

\textbf{HazeRD:} The HazeRD dataset is a simulated hazy dataset comprising 15 real-world outdoor scenes. For each scene, multiple variants were generated by adjusting different parameters, resulting in five different weather condition images for each scene. Therefore, this dataset includes a total of 75 hazy images and their original 15 real images.

\subsection{Implementation Details}
During training, our experimental settings were consistent with the baseline model DM2F~\cite{deng2019deep}. We initialized the basic CNN parameters using the ResNeXt~\cite{Xie2016AggregatedRT} pre-trained on ImageNet~\cite{5206848}, while other parameters were initialized with Gaussian random noise. We used the Adam optimizer~\cite{2014Adam}, setting the number of iterations to 40,000 for the RESIDE dataset and 20,000 for the O-HAZE dataset. The learning rate was adjusted using the poly strategy~\cite{Liu2015ParseNetLW}, with an initial learning rate of 0.0002 and a decay power of 0.9. We randomly cropped 256×256 image patches from the entire training images for training, with a batch size of 16. The implementation was done in the PyTorch framework, and training and inference tests were conducted on a machine with a single NVIDIA 3090 GPU, completing model training in about 5 hours.

\subsection{Ablation Study}

\begin{table}[ht]
\centering
\caption{Average PSNR and SSIM values in ablation study}
\begin{tabular}{lcc}
\toprule
\multirow{2}{*}{Method} & \multicolumn{2}{c}{RESIDE\cite{Ren-CVPR-2018}} \\
\cmidrule(r){2-3}
 & PSNR & SSIM \\
\midrule
\midrule
FD - AS & 33.87 & 0.9715 \\
FD - $J_1$ & 35.33 & \textbf{0.9814} \\
FD - $J_2$ & 35.05 & 0.9790 \\
FD - $J_3$ & 34.76 & 0.9786 \\
FD - $J_4$ (CL2S, Ours) & \textbf{35.36} & 0.9808 \\
FD - $J_5$ (DM2F, Baseline) & 34.99 & 0.9804 \\
FD - $J_{1, 4}$ & 35.13 & 0.9809 \\
FDNet & 35.25 & 0.9798 \\
\bottomrule
\end{tabular}
\end{table}

We first questioned the ablation method of the DM2F network, pointing out that it only isolated the basic effectiveness of each component without discussing the impact on overall performance when any component is removed, which is a shortcoming in the original study. To this end, we conducted a series of systematic ablation experiments to comprehensively review the indispensability of each functional component in DM2F. Based on this, we constructed an enhanced model—the Full Elementary Function Dehazing Network (FDNet), which integrates five basic operator models, including logarithmic and sine functions, forming a complete framework.

To explore the role of each component in depth, we designed six variant models based on FDNet and conducted comprehensive evaluations on the RESIDE dataset~\cite{Ren-CVPR-2018}. The first variant (labeled "FD - AS") was a control experiment by removing the atmospheric scattering model. Then, by individually removing $J_1$ (labeled "FD - $J_1$"), $J_2$ (labeled "FD - $J_2$"), $J_3$ (labeled "FD - $J_3$"), $J_4$ (corresponding directly to the CL2S model, labeled "FD - $J_4$"), and $J_5$ (labeled "FD - $J_5$"), we constructed another four models. Finally, we established a special model variant (labeled "FD - $J_{1, 4}$") by simultaneously removing $J_1$ and $J_4$ to examine whether reducing the number of elementary functions could maintain the model's performance boundary, thus deepening the understanding of network structure optimization. Based on the above meticulously designed experimental analysis, we have reason to believe that the four basic operators adopted in the current model—linear operations (addition and multiplication), exponential functions, and sine functions—constitute the most optimized combination of elementary functions for efficiently simulating haze phenomena. This configuration not only highlights the advantage of function diversity in capturing hazy image characteristics but also demonstrates that maintaining this balanced combination of four elementary functions is crucial for achieving highly realistic dehazing effects.

\section{Conclusion and Future Work}

This study presents an innovative dehazing network architecture, CL2S (Change Logarithmic to Sinusoidal function), based on an in-depth analysis of the challenges and limitations of existing image dehazing techniques. The network effectively upgrades the DM2F network by introducing trigonometric functions, specifically the sinusoidal model, in place of the original logarithmic model. Experimental results demonstrate that CL2S outperforms baseline methods on multiple standard datasets, proving the feasibility and superiority of incorporating trigonometric functions into dehazing models. Through detailed ablation experiments, we not only verified the importance of each elementary function model but also gained a deeper understanding of their interactions, providing new insights for constructing efficient dehazing models. Notably, the introduction of the sinusoidal model enriches the model's capability to handle complex haze effects, enhancing its robustness and generalization performance, and showcasing excellent dehazing potential in complex environments.

Despite the positive results achieved by the CL2S network in image dehazing tasks, several avenues for future research remain worth exploring. (1) Further investigation into the combination of different elementary functions and their parameter optimization strategies may uncover more efficient and generalizable dehazing models. (2) Exploring unsupervised or semi-supervised learning strategies could reduce the reliance on large-scale hazy-clear image pairs, enhancing the model's adaptability in real-world scenarios. (3) Combining physical models with deep learning approaches could leverage physical principles to constrain model learning, while simultaneously improving the physical realism of the dehazing results.

{
    \small
    \bibliographystyle{ieeenat_fullname}
    \bibliography{main}
}


\end{document}